\documentclass{article}

\PassOptionsToPackage{numbers,compress}{natbib}
\usepackage[preprint]{neurips_2026}
\workshoptitle{Robot Learning with World Models (RLWM)}

\usepackage[utf8]{inputenc}
\usepackage[T1]{fontenc}
\usepackage{hyperref}
\usepackage{url}
\usepackage{booktabs}
\usepackage{amsmath,amsfonts,amssymb}
\usepackage{nicefrac}
\usepackage{microtype}
\usepackage{xcolor}
\usepackage{graphicx}
\usepackage{tabularx}
\usepackage{array}
\title{Beyond Task Success: Stage-Wise Reliability of World Model Planning under Sensing Degradation}
\author{
Geonmyeong Lee$^{1}$, Byoung-Tak Zhang$^{1}$\\
Seoul National University$^{1}$\\
\{gmlee, btzhang\}@bi.snu.ac.kr
}

\begin{document}
\maketitle

\begin{abstract}
In world model planning, sensing inputs pass through an encoder and predictor before affecting planner decisions, so final task success alone cannot reveal where sensing disturbances attenuate or persist in the pipeline.
We apply 10 visual and temporal sensing degradations to a world model planner and track their effects across representation, future prediction, planner preference, and physical outcome using paired evaluation on the same 50 tasks.
The relative impact of degradations was not preserved across stages: large representation shifts could attenuate downstream, while smaller initial shifts could persist to the outcome, and internal-response ordering did not directly match physical-outcome ordering.
Temporal degradations also showed distinct patterns: even with similar overall changes in observation history, responses differed substantially with the location of corrupted information and the planner's actual exposure.
This non-uniform stage-wise response was also observed in secondary evaluations with another manipulation task and a different world model.
Stage-wise diagnosis can therefore identify where sensing disturbances attenuate or persist and help prioritize subsequent model verification and sensing mitigation.
\end{abstract}

\section{Introduction}\label{sec:introduction}

World model planning is increasingly being combined with Vision-Language-Action (VLA) models and applied to real-robot manipulation
\cite{DINO-WM,WORLDVLA,WM-VLA}.
Because these systems predict futures and choose actions from sensing information including visual observations, sensing degradation in real environments is a key reliability concern.
Camera-based sensing can suffer not only appearance and visibility degradations such as illumination changes, blur, and occlusion, but also temporal degradation such as frame loss and observation delay.

Prior work in vision, robot learning, and world models has studied robustness to sensing perturbations
\cite{LIBERO-PLUS,INVARIANT-WM},
while recent studies suggest that prediction quality alone may not explain downstream planning reliability and motivate internal diagnostics
\cite{CONTROL-THEORY,ACPC,VISCORE}.
Can sensing robustness of a multi-stage world model planner therefore be assessed by final task outcome alone?
Task outcome reveals success or failure but not where a disturbance attenuates or how far it persists through the pipeline.
A stage-wise view can instead narrow the failures and functional stages that subsequent model verification and sensing mitigation should target.

\begin{figure*}[t]
\centering
\includegraphics[width=\textwidth]{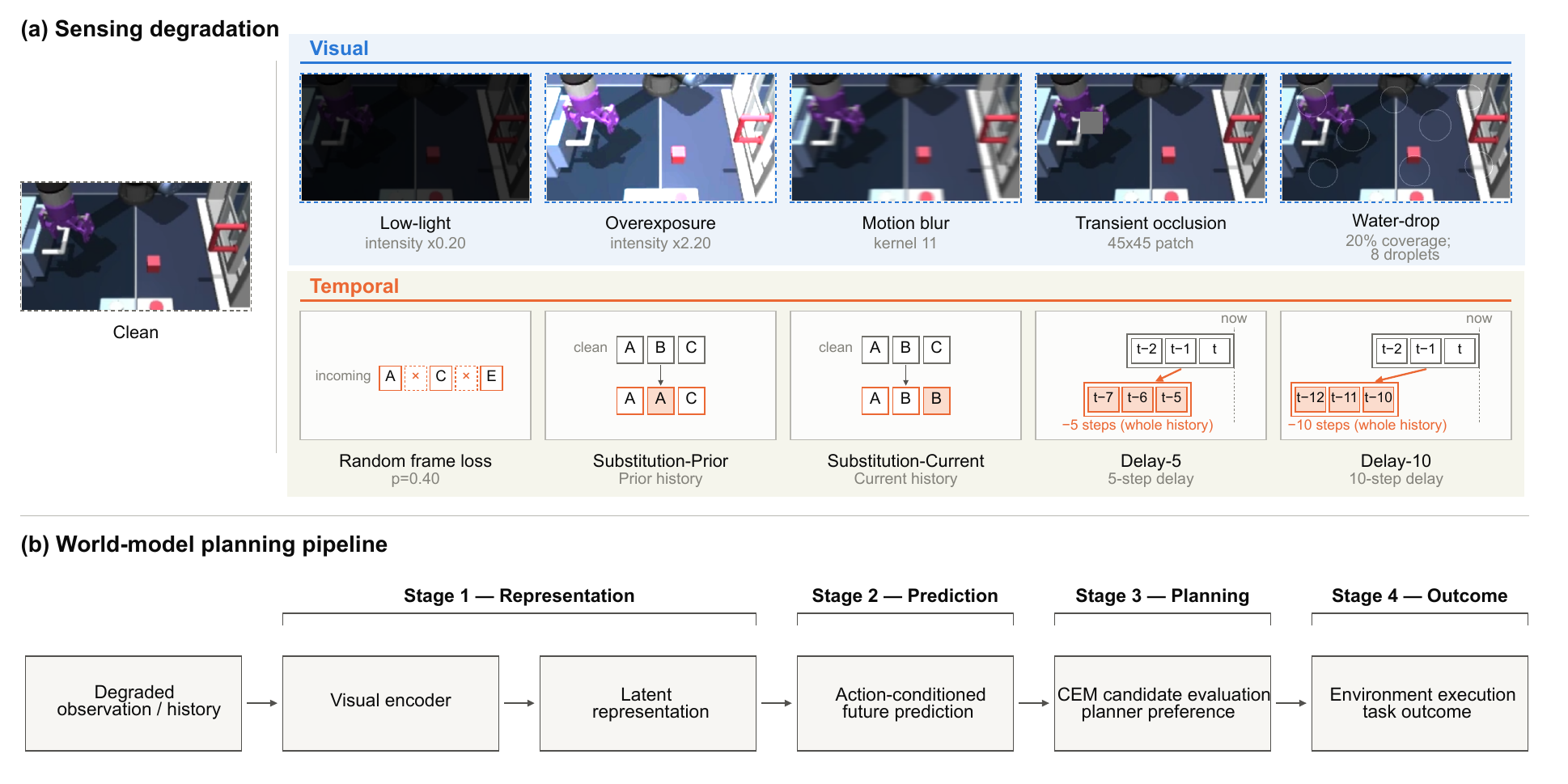}
\caption{
(a) Visual and temporal sensing conditions in the primary evaluation.
(b) Stage 1--4 track changes in representation, prediction, planning, and task outcome. Degradation is applied to the observation/history while the goal observation remains clean.
}
\label{fig:pipeline}
\end{figure*}

To study this question, we formulate diverse sensing conditions as controlled shifts applied to the same world model planner and compare each against a paired clean condition.
We evaluate downstream changes under 10 visual and temporal sensing degradations, with temporal degradation separated into random frame loss, history substitution, and fixed delay.
We further examine stage-wise responses on another manipulation task and a separate world model as secondary evaluations.
Figure~\ref{fig:pipeline} summarizes the sensing conditions and stage-wise evaluation framework.
Our main contributions are as follows.

\textbf{Stage-wise sensing analysis.}
We construct a paired stage-wise evaluation that tracks each sensing disturbance from representation through future prediction, planner preference, and task outcome, enabling comparison of downstream propagation across degradations.

\textbf{Degradation-specific propagation.}
The relative impact of sensing degradations did not remain ordered across the pipeline.
Some disturbances attenuated strongly after representation, whereas others persisted through prediction and planning, and large planner responses did not necessarily correspond to lower success-rate point estimates.
Thus, neither a single internal metric nor final outcome alone fully characterizes sensing degradation.

\textbf{Temporal information structure.}
Random frame loss, history substitution, and fixed delay produced distinct propagation profiles, and similar history-level changes led to very different prediction and planning responses depending on where information was corrupted.
Temporal degradation therefore requires considering information position, freshness, and the planner's actual exposure in addition to aggregate history change or nominal corruption.

\section{Related Work}\label{sec:related-work}

\paragraph{Sensing degradation and robustness.}
Robustness to perturbations such as lighting, viewpoint, background, and sensor noise has been systematically evaluated in robot learning and VLA systems, while robust planning under visual variation has also been studied in world models
\cite{LIBERO-PLUS,INVARIANT-WM,STABLE-WM}.
For temporal sensing, prior work has examined the effects of missing observations and observation delay on control and their mitigation
\cite{DELAY-RL}.
These studies motivate the sensing variations considered here; our focus is not the mechanism or mitigation of individual degradations, but how major sensing disturbances persist through downstream stages of world model planning.

\paragraph{Stage-wise diagnostics and error propagation.}
Recent world model studies show that prediction quality and downstream control performance need not align
\cite{CONTROL-THEORY}.
ACPC diagnoses action-conditioned predictive consistency under primarily noise-based visual perturbations from a bisimulation perspective
\cite{ACPC},
while VIScore proposes a planning-relevant model-quality diagnostic spanning the encoder, predictor, and planner
\cite{VISCORE}.
These works motivate examining world model planning across functional stages.
We connect this perspective to sensing degradation and track how diverse disturbances change as they pass through the same pipeline.

\section{Sensing Degradation and Stage-Wise Evaluation}\label{sec:method}\label{sensing-degradation-and-stage-wise-evaluation}

\subsection{Sensing degradation conditions}\label{sensing-degradations}

We use 10 synthetic degradations that simplify sensing problems that may occur in a robot's visual observations.
Table~\ref{tab:sensing_conditions} specifies each condition.
Appearance/visibility conditions alter image brightness, sharpness, or visible regions, whereas temporal conditions introduce missing, substituted, or delayed observations into the recent history used by the planner.

\paragraph{Appearance and visibility degradation.}
Low-light and overexposure decrease and increase global image intensity, respectively, with clipping in bright regions under overexposure.
Motion blur smears object boundaries and textures to mimic camera or robot motion.
Transient occlusion temporarily masks a random \(45\!\times\!45\) image patch.
The water-drop condition approximates droplets on the lens by applying local blur and refraction to multiple fixed droplet regions throughout an episode.

\paragraph{Temporal degradation: random frame loss, history substitution, and observation delay.}
Temporal conditions modify recent observations through stochastic missing updates, substitution at a specific history position, or fixed observation delay.
Random frame loss drops each new frame with 40\% probability.
To further test planner response to selective temporal-information recovery, we compare all-delay history, history with only the latest observation restored, and clean history at the same physical state.

\begin{table*}[t]
\centering
\caption{Sensing degradation conditions used in the primary evaluation.}
\label{tab:sensing_conditions}
\small
\setlength{\tabcolsep}{5pt}
\begin{tabularx}{\textwidth}{@{}ll l >{\raggedright\arraybackslash}X@{}}
\toprule
Family & Condition & Parameter & Operation \\
\midrule
Visual & Low-light & gain 0.20
& decrease image intensity \\
Visual & Overexposure & factor 2.20
& increase image intensity with clipping \\
Visual & Motion blur & kernel 11
& apply directional blur \\
Visual & Transient occlusion & \(45\times45\) patch
& temporarily mask a local visual region \\
Visual & Water-drop & 20\% coverage, 8 droplets
& apply local blur and refraction in droplet regions \\
\midrule
Temporal & Random frame loss & \(p=0.40\)
& reuse the last valid frame when an update is dropped \\
Temporal & Substitution-Prior & \([A,B,C]\rightarrow[A,A,C]\)
& replace a prior history observation with an older one \\
Temporal & Substitution-Current & \([A,B,C]\rightarrow[A,B,B]\)
& replace the latest observation with the previous one \\
Temporal & Delay-5 & 5-step delay
& shift the observation history to the stream 5 environment steps earlier \\
Temporal & Delay-10 & 10-step delay
& shift the observation history to the stream 10 environment steps earlier \\
\bottomrule
\end{tabularx}
\end{table*}

\subsection{Stage-wise analysis}\label{stage-wise-analysis}

Figure~\ref{fig:pipeline} maps four analysis stages to the functional flow of world model planning.
Stage 1 measures changes in observation representation; Stage 2 measures how much of that change remains in future prediction under the same action; Stage 3 measures changes in planner action preference; and Stage 4 measures changes in final task outcome.
Although the CEM planner couples future prediction and candidate-action evaluation, we separate Stages 2 and 3 for diagnosis.

\paragraph{Stage 1 --- Representation.}\label{stage-1-representation}
We feed clean and degraded visual inputs to the frozen visual encoder and measure the distance between their latent representations.
Because latent-distance scales differ across models, we normalize by the representation change from one normal step along a clean trajectory.
For temporal conditions, we additionally measure the current-observation distance \(D_{0,\mathrm{current}}\) to separate whole-history shift from change in the latest observation itself.
Thus, Stage 1 measures the degradation-induced representation shift relative to a model's normal one-step change.

\paragraph{Stage 2 --- Prediction.}\label{stage-2-prediction}
Stage 2 feeds the same action sequence to clean and degraded contexts and measures the difference between predicted future representations, testing whether observation-induced discrepancy attenuates or persists through prediction.
For each degradation, we use the action sequence selected by the degraded planner and inject that same sequence into both contexts, isolating context-induced prediction differences from differences caused by action selection.
Because the initial discrepancy \(D_{\mathrm{context}}\) varies across degradations, absolute \(D_5\) can overstate conditions that begin with larger distortions.
We therefore use the relative residual \(D_5/D_{\mathrm{context}}\) as the primary metric: values near zero indicate attenuation during rollout, whereas larger values indicate stronger persistence into future prediction.

\paragraph{Stage 3 --- Planning.}\label{stage-3-planning}
Stage 3 evaluates how sensing degradation changes planner action preference.
Independent clean and degraded CEM runs would sample different action candidates, so we fix the same 300 candidate sequences generated at CEM iteration 0 and evaluate them under both observation contexts.
Candidates are scored by latent-distance cost to the clean goal; the main metrics are the degraded-context rank of the clean winner and the number and fraction of episodes in which it remains rank 1 (same-top).
Because CEM may choose different valid paths, this directly measures preference changes over a shared candidate pool rather than only the distance between final action vectors.

\paragraph{Stage 4 --- Outcome.}\label{stage-4-outcome}
Stage 4 independently runs the planner with clean and degraded observations, executes the action actually selected in each condition, and compares task outcomes on the same start--goal task.
Success follows each environment's evaluator-defined goal criterion within the episode budget.
We report each condition's success rate, the paired success-rate difference \(\Delta\mathrm{SR}\), and the 95\% Tango score confidence interval for paired binary proportions.

\subsection{Evaluation setup}\label{evaluation-setup}

\paragraph{Primary Scene evaluation.}
We use DINO-WM \cite{DINO-WM} as the primary model because its frozen visual representation, learned action-conditioned prediction, and CEM-based planning expose functional boundaries suitable for stage-wise analysis.
The primary environment is the OGBench OGBScene Drawer manipulation task.
We use goal offset 20 as an operating point that avoids an overly easy single-shot setting and exposes replanning and temporal-degradation effects.
The final Scene evaluation uses the same fixed 50 replan-eligible start--goal tasks, pairing clean and degraded conditions on each task.

\paragraph{Secondary Cube and LeWM evaluation.}
To test whether the Scene findings extend to another manipulation task and world model architecture, we perform a secondary stage-wise evaluation on OGB-Cube.
Whereas OGBScene Drawer changes the state of an articulated drawer, OGB-Cube relocates an independent cube to a specified 3-D target position.
On the same Cube task, we evaluate DINO-WM and LeWM \cite{LEWM}, a JEPA-style world model with different representations and training, using low-light, motion blur, and Delay-10 as representative appearance/temporal contrasts.
To compare how degradation distorts otherwise successful behavior, each model is evaluated only on tasks it solves under clean sensing.

\begin{figure*}[t]
\centering
\includegraphics[width=0.9\columnwidth]{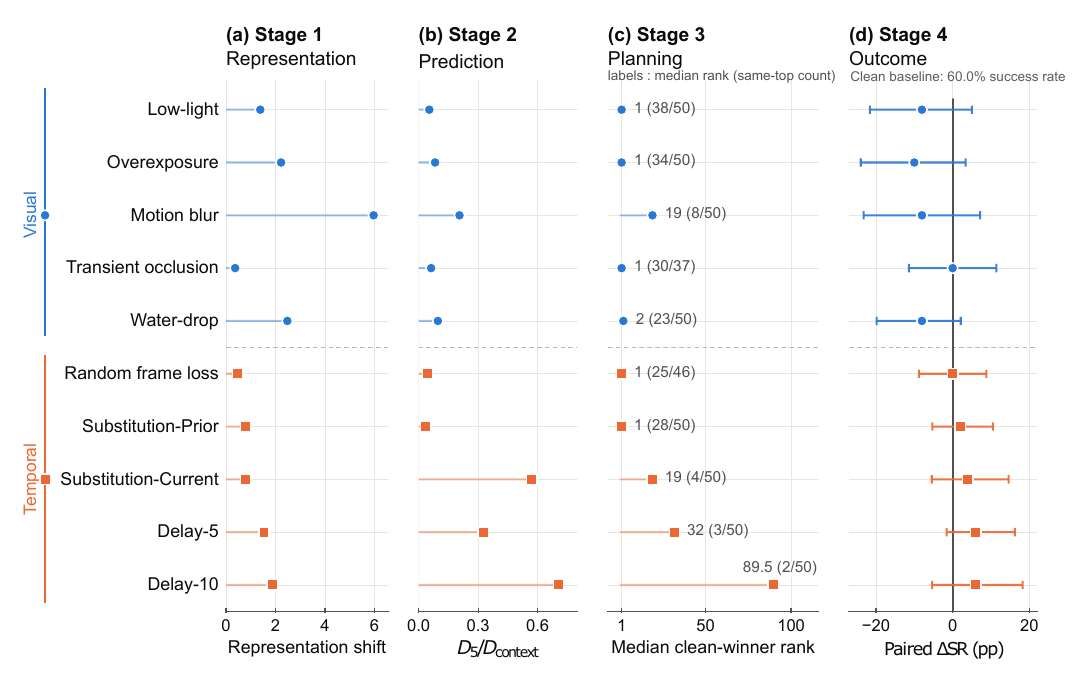}
\caption{Stage-wise sensing responses in the primary Scene evaluation.}
\label{fig:main_results}
\end{figure*}

\section{Results}\label{sec:results}\label{results}

Figure~\ref{fig:main_results} summarizes the effects of all 10 sensing degradations across Stages 1--4 in the primary Scene evaluation.

\subsection{Sensing degradations show distinct propagation profiles}
\label{different-sensing-degradations-show-distinct-propagation-patterns-across-the-pipeline}

\paragraph{Internal-stage propagation: representation changes did not preserve their ordering downstream.}
Across Stages 1--3 in Figure~\ref{fig:main_results}, degradation effects did not propagate uniformly.
Low-light and overexposure produced Stage 1 representation shifts of 1.384 and 2.224, yet their Stage 2 \(D_5/D_{\mathrm{context}}\) values fell to 0.054 and 0.083 and their Stage 3 median ranks remained 1, indicating strong downstream attenuation.
By contrast, Substitution-Current produced a smaller Stage 1 shift of 0.792 but retained 0.571 at Stage 2 and moved the median clean-winner rank to 19, showing stronger persistence downstream.

Even conditions with similar Stage 2 and aggregate Stage 3 statistics differed in same-top reordering.
Overexposure and water-drop had similar Stage 2 values (0.083 and 0.097) and median clean-winner ranks (1 and 2), but same-top was 34/50 versus 23/50, and the worst clean-winner rank was 8 versus 26.

\paragraph{Internal-to-outcome decoupling: large internal changes did not necessarily imply large task-performance drops.}
Low-light and overexposure strongly attenuated at Stage 2 and both retained median rank 1 at Stage 3, yet Stage 4 success fell from the clean baseline of 60.0\% to 52.0\% and 50.0\%, giving \(\Delta\mathrm{SR}=-8.0\) and \(-10.0\).
Conversely, Delay-5 and Delay-10 showed high Stage 2 residuals of 0.326 and 0.705 and large Stage 3 median-rank shifts to 32 and 89.5, yet Stage 4 success increased to 66.0\% for both.
Motion blur produced the largest Stage 1 representation shift but reached 52.0\% success (\(\Delta\mathrm{SR}=-8.0\)), a smaller point decrease than overexposure.

The ordering of internal-stage diagnostics therefore did not match Stage 4 point estimates.
This mismatch does not imply that internal changes vanished during rollout: internal change and task-relevant information are different quantities, distinct action sequences can reach the same successful outcome, and closed-loop replanning may reduce early decision differences.
Our experiments do not causally separate these possibilities.
Moreover, all paired 95\% confidence intervals for primary conditions included zero, so Stage 4 point estimates should not be interpreted as evidence of superiority or improvement across conditions, nor as evidence of no effect or equivalence.

\paragraph{Appearance--temporal Stage 4 contrast: the two families showed opposite mean $\Delta\mathrm{SR}$ directions.}
Appearance conditions generally reduced Stage 4 success, whereas some temporal conditions increased it.
Clean-success-to-failure transitions differed markedly between appearance and temporal conditions (32/150 vs. 4/150), while clean-failure-to-success transitions were similar (15/100 vs. 13/100), suggesting that positive temporal \(\Delta\mathrm{SR}\) mainly reflected fewer down-flips.
We therefore further examined this directional Stage 4 contrast.

Among tasks successful under clean sensing, terminal-error increases of at least \(40\,\mathrm{mm}\) occurred in 32/150 appearance cases but only 3/150 temporal cases; 35 of 36 clean-success-to-failure transitions were included in these large-error cases.
Yet any error increase occurred at similar frequencies (85/150 appearance vs. 79/150 temporal), indicating that the Stage 4 degradation under appearance conditions was concentrated in more frequent large-error cases rather than uniform worsening across tasks.
The \(40\,\mathrm{mm}\) threshold is a post-hoc diagnostic for changes larger than one task-success tolerance width, not a new success criterion.
Large-error cases did not share one physical failure pattern, and they occurred in 2 of 11 tasks whose first plan was clean and only later encountered occlusion, so initial-planning corruption was not necessary.

These results show that sensing effects are not simply transmitted proportionally through the pipeline and motivate stage-wise evaluation.
Final success rate alone cannot distinguish whether a disturbance already attenuated during prediction, altered planner preference, or left task outcome unchanged despite large internal shifts.
Stage-wise diagnosis is therefore better suited to locating how far a sensing disturbance propagates and narrowing targets for robustness interventions than to directly predicting Stage 4 from any single internal metric.

\subsection{Temporal degradations show distinct propagation profiles}
\label{temporal-results}

\begin{figure*}[!t]
\centering
\includegraphics[width=0.9\columnwidth]{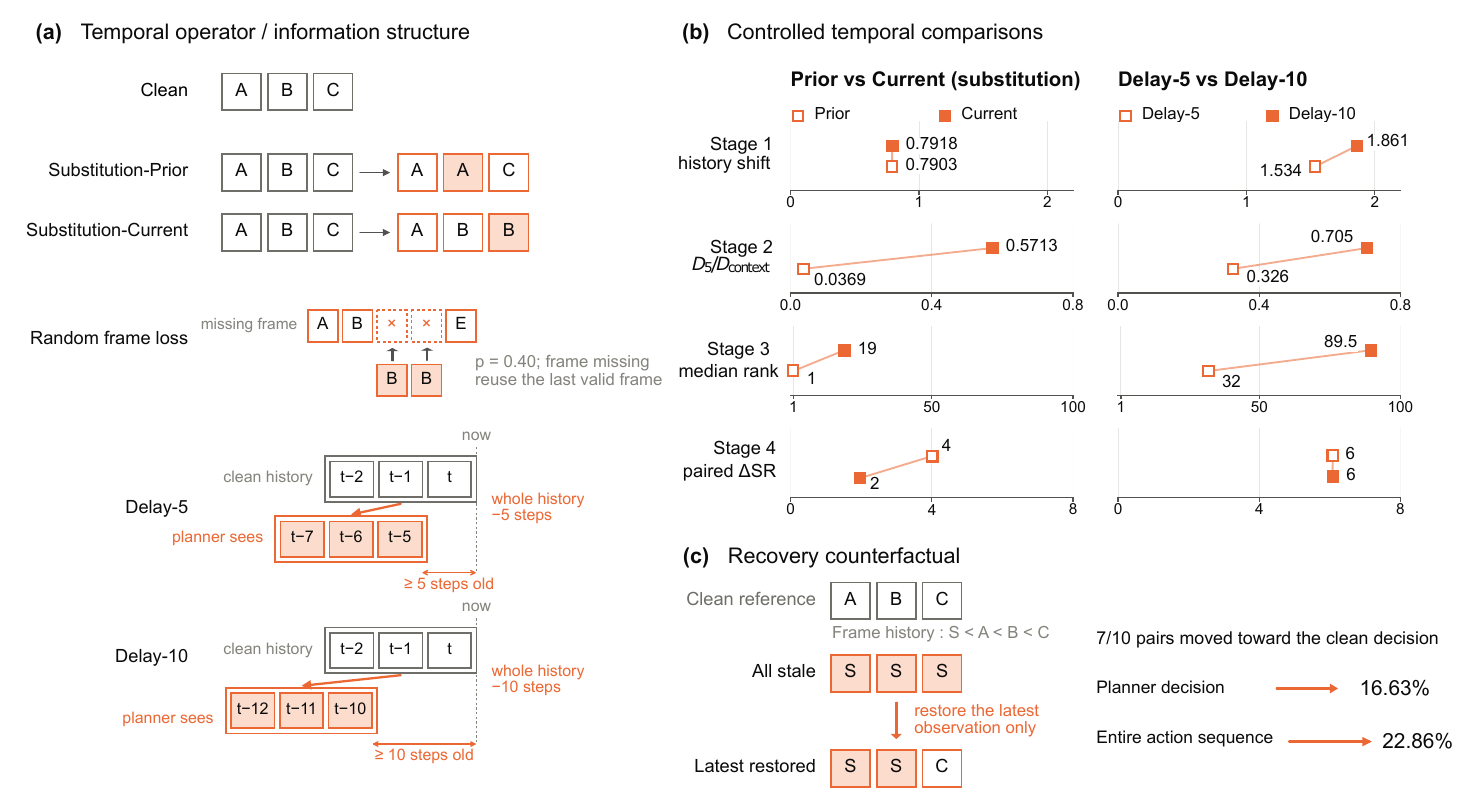}
\caption{
(a) Temporal conditions.
(b) Stage-wise comparison of Prior--Current and Delay-5--Delay-10.
(c) Supporting DEV counterfactual that restores only the latest observation at the same physical state.
}
\label{fig:temporal}
\end{figure*}

\paragraph{Random missing updates, history substitution, and delay produced different profiles.}
Temporal sensing degradations did not share one propagation pattern.
Random frame loss and Substitution-Prior had Stage 1 values of 0.463 and 0.790, but small Stage 2 values of 0.045 and 0.037 and median clean-winner rank 1 at Stage 3.
Substitution-Current had a nearly identical Stage 1 value of 0.792, yet retained 0.571 at Stage 2 and moved the median clean-winner rank to 19.
Thus, temporal operators differed not only in effect magnitude but also in propagation according to actual exposure and the position of corrupted information in history.

\paragraph{Actual exposure and history position strongly shaped temporal effects.}
Two important factors were whether the planner actually consumed a corrupted observation and where that corruption appeared in history, particularly whether the latest observation remained fresh.
Because random frame loss is stochastic, the history consumed by the planner varied across tasks.
Dropped frames entered the actual history in 46 of 50 tasks, but the latest observation itself was dropped in only 21.
Correspondingly, random frame loss caused relatively small changes across Stages 1--3, showing that temporal corruption should be characterized by actual planner exposure as well as the nominal condition.

History position was clearer in the contrast between Substitution-Prior and Substitution-Current.
Despite similar representation-change magnitudes, prediction and planner preference differed substantially depending on which observation became stale, with latest-observation freshness closely associated with planner sensitivity.
However, Substitution-Prior still displaced the clean-best action from rank 1 in 22/50 tasks, so the current observation should not be interpreted as the only important history element or as always dominating planner decisions.
Increasing observation delay from 5 to 10 steps also increased changes at every stage, indicating larger effects across representation, prediction, and planning.

\paragraph{Recovery: restoring the recent observation partially moved planner decisions toward clean.}
When only the latest observation was restored to a fresh value, planner decisions moved toward the clean decision in 7 of 10 pairs; the median distance to the clean decision decreased by 16.63\% for the first action and 22.86\% for the full action sequence.
This provides supporting evidence that selectively restoring temporal information identified by stage-wise diagnosis can partially recover downstream planner behavior.
However, this is a DEV counterfactual that changes history at the same physical state; it does not establish recovery of the subsequent robot trajectory or task success after fresh observations return.

\subsection{Stage-wise evidence across another task and model}\label{secondary-evidence-across-environments-and-models}

To test whether the primary Scene findings were specific to one task and model, we evaluate DINO-WM and LeWM on the same OGB-Cube task as a secondary extension.
Table~\ref{tab:cube_lewm} summarizes Stages 1--4 for low-light, motion blur, and Delay-10.
DINO-WM again showed different stage-wise orderings on Cube, while LeWM exhibited different appearance sensitivity from DINO-WM.
Although the most disruptive appearance degradation depended on the model, stage-wise heterogeneity---the failure of one degradation to retain the same relative impact across all stages---reappeared in the second world model.
Delay effects also persisted into prediction and planner preference in both models.

\begin{table*}[t]
\centering
\caption{Stage-wise results on the secondary Cube evaluation. Stage 1 is normalized within each model; parentheses for Delay-10 in Stage 4 indicate the planner-exposed subset.}
\label{tab:cube_lewm}
\small
\setlength{\tabcolsep}{5pt}
\begin{tabular}{@{}llrrrrrl@{}}
\toprule
& & Stage 1 & Stage 2 & \multicolumn{2}{c}{Stage 3} & Stage 4 \\
\cmidrule(lr){5-6}
Model & Condition
& Repr.
& \(D_5/D_{\mathrm{context}}\)
& Median rank
& Same-top
& Success \\
\midrule
DINO-WM & Low-light
& 2.46 & 0.228 & 2 & 6/17 & 8/17 \\
DINO-WM & Motion blur
& 4.69 & 0.104 & 3 & 7/17 & 14/17 \\
DINO-WM & Delay-10
& 1.88 & 0.947 & 93.5 & 0/12 & 14/17 (9/12) \\
\addlinespace[2pt]
LeWM & Low-light
& 7.13 & 0.846 & 93 & 2/15 & 2/15 \\
LeWM & Motion blur
& 0.142 & 0.954 & 2 & 7/15 & 11/15 \\
LeWM & Delay-10
& 3.95 & 0.990 & 157 & 0/5 & 12/15 (2/5) \\
\bottomrule
\end{tabular}
\end{table*}

\section{Conclusion}\label{sec:conclusion}

We traced the effects of visual and temporal sensing degradation across representation, future prediction, planner preference, and task outcome in world model planning.
Sensing disturbances did not simply pass through the pipeline; instead, each condition showed distinct attenuation and persistence patterns.
For temporal degradation, downstream response depended not only on overall history change but also on information position, freshness, and the planner's actual exposure.
Stage-wise diagnosis can therefore distinguish where disturbances attenuate or persist and help narrow the failures and functional stages targeted by model verification and sensing mitigation.

Because this study uses controlled synthetic degradations and a limited set of world model planning settings, future work should test the scope of these patterns across broader models, tasks, and real sensing conditions.
A further direction is to directly evaluate whether targeted interventions on diagnosed stages or sensing failures recover downstream planning and physical outcomes.

\bibliographystyle{plainnat}
\bibliography{references}


\end{document}